\pdfoutput=1

\documentclass[11pt]{article}

\usepackage[]{EMNLP2022}

\usepackage{times}
\usepackage{latexsym}
\usepackage{graphicx}
\usepackage{booktabs}
\usepackage{float}
\usepackage{comment}
\usepackage{xfrac}
\usepackage[T1]{fontenc}

\usepackage[utf8]{inputenc}

\usepackage{microtype}

\usepackage{inconsolata}

\title{ JamPatoisNLI: A Jamaican Patois Natural Language Inference Dataset}

\newcommand{\AnD}{\hskip 2em plus 1fil minus 0.5em}
 \author{Ruth-Ann Armstrong \AnD John Hewitt \AnD Christopher Manning \\
 Department of Computer Science\\
 Stanford University \\
 \{\texttt{ruthanna,johnhew,manning}\}\texttt{@cs.stanford.edu}}

\begin{document}
\maketitle

\begin{abstract}

JamPatoisNLI provides the first dataset for natural language inference in a creole language, Jamaican Patois.
Many of the most-spoken low-resource languages are creoles. These languages  commonly have a lexicon derived from a major world language and a distinctive grammar reflecting the languages of the original speakers and the process of language birth by creolization. This gives them a distinctive place in exploring the effectiveness of transfer from large monolingual or multilingual pretrained models. While our work, along with previous work, shows that transfer from these models to low-resource languages that are unrelated to languages in their training set is not very effective, we would expect stronger results from transfer to creoles. Indeed, our experiments show considerably better results from few-shot learning of JamPatoisNLI than for such unrelated languages, and help us begin to understand how the unique relationship between creoles and their high-resource base languages affect cross-lingual transfer. JamPatoisNLI, which consists of naturally-occurring premises and expert-written hypotheses, is a step towards steering research into a traditionally underserved language and a useful benchmark for understanding cross-lingual NLP\@. 
\end{abstract}

\section{Introduction}

The extensive progress that has been made in NLP research in recent years has largely been constrained to around 20 of the 7000 languages spoken around the world \cite{lrls}. Creole languages, which emerge as a result of contact between speakers of different vernaculars, are even further underexplored \cite{creolegap}.

 This work contributes to addressing this gap. We present JamPatoisNLI, the first natural language inference dataset in Jamaican Patois, which is an English-based creole spoken in the Caribbean.  Additionally, to our knowledge, no other natural language inference corpus exists for any other creole language. \\
 \begin{figure}[t]

\includegraphics[width=0.48\textwidth]{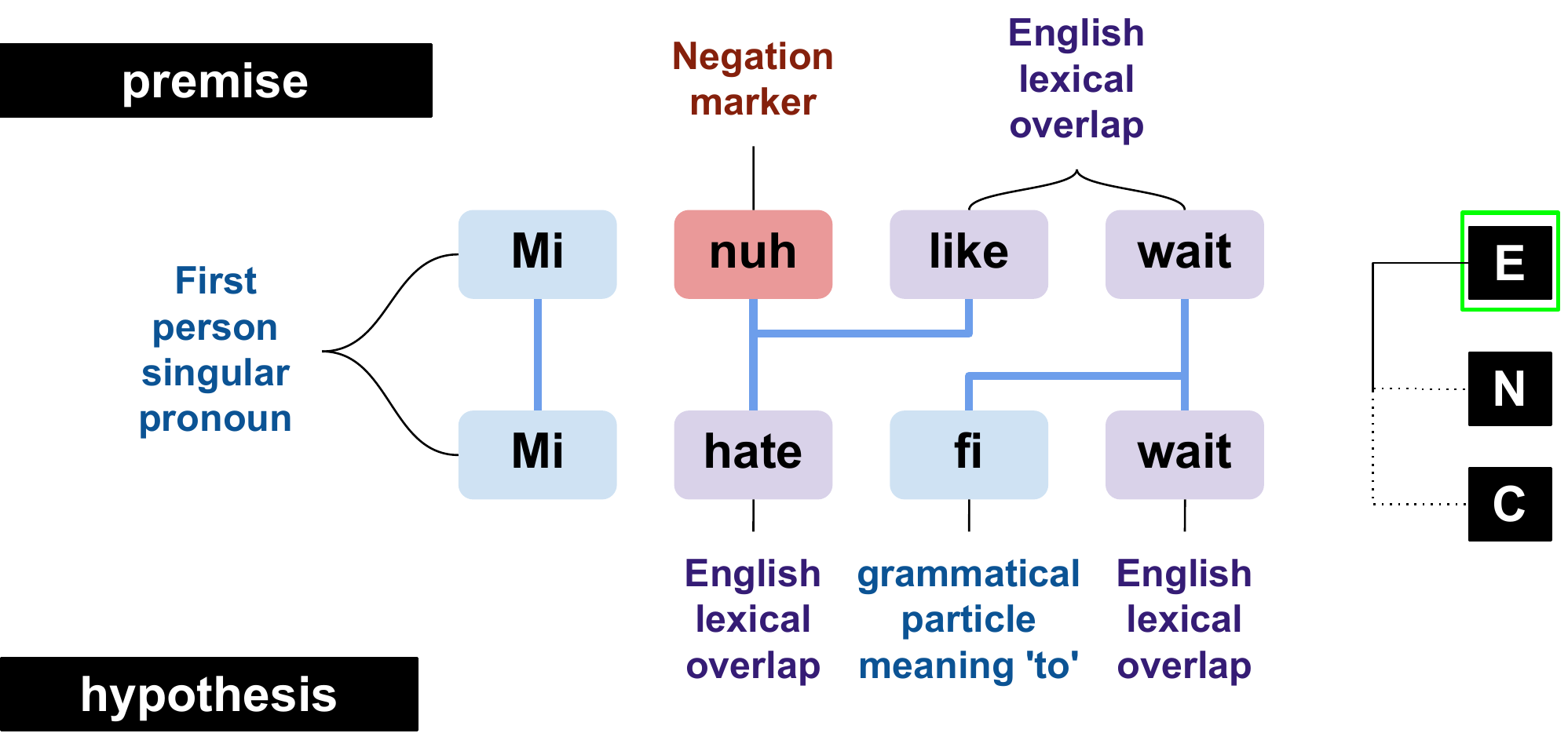}

\caption{\label{figure_header} Linguistic features relevant for textual entailment classification for Jamaican Patois and lexical overlap with English.}
\end{figure}
 Jamaican Patois is one of over 100 creole languages spoken by millions of inhabitants of different regions across the world, including Africa, the Caribbean, the Americas, islands in the Indian Ocean and the Pacific Ocean (including Australia and the Philippines) and South Asia \cite{creolenumbers, creoleregions}. Though there has been a recent spike in interest in work on  low-resource languages in the NLP community \cite{lrl3uzbek, lrl10qa, americasnli, lrl4hausa, lrl2bengali, lrl1parallel, lrl8synthetic, lrl9augmented, lrl6ner, lrl11nlg}, creoles in particular are extremely under-explored in spite of the prevalence of their usage globally \cite{creolegap}. Working more with this class of languages is an important step in ensuring that the benefits of NLP technology are more equitably distributed globally.
 
 Additionally, the class of creole languages is a uniquely interesting point of study within the space of multilingual NLP. Though creoles like Jamaican Patois have distinct morphosyntactic features, they often share significant lexical overlap with the high-resource base languages from which they are derived. This makes it possible to study cross-lingual transfer between high-resource and low-resource languages that are distinct, but share similar lexicons. In particular, JamPatoisNLI provides a benchmark for NLP researchers working to understand cross-lingual transfer to languages outside the training data of large pretrained multilingual models. Creole languages like Jamaican Patois have the unique property of being outside the pretraining data of these models, yet highly related to their base languages, which are present in the datasets used to train the models.\footnote{In large web scrapes, there likely is some Jamaican Patois language in the resulting text, but it is not, e.g., one of the languages with a Wikipedia large enough to be included in Multilingual BERT.}

JamPatoisNLI was constructed using both naturally occurring and newly constructed utterances of Jamaican Patois rather than through translation. This mitigates the problem of skewed cross-lingual transfer results which arises when the test dataset consists of translated examples but the training dataset does not \cite{artifacts}. This also enhances the \textit{ecological validity} \cite{ecovalid} of the dataset, as it is grounded in real world usage of the language and is thus a more relevant, realistic benchmark. These two features mean that work done with the dataset will be particularly useful for moving towards developing technologies for  speakers of the language.

We run studies on JamPatoisNLI transferring from monolingual English BERT, multilingual BERT, monolingual English RoBERTa and multilingual XLM-RoBERTa, finetuned on the Multi-NLI dataset, in zero-shot and few-shot settings. We find that monolingual English RoBERTa (76.50\%) and multilingual XLM-RoBERTa (75.17\%) achieve similar accuracies when we use the entire few-shot JamPatoisNLI training dataset with 250 examples for further finetuning.
We also find that the monolingual English BERT model (66.17 \%) and the multilingual BERT model (65.33 \%), achieve similar accuracies when we use the entire few-shot JamPatoisNLI training dataset. In our experiments, the RoBERTa-based models strongly outperform the BERT-based models.  Additionally, we find that few-shot performance on JamPatoisNLI increases much faster (with respect to the number of few-shot training examples) than on languages in AmericasNLI, which have no strong connection to a high-resource language \cite{americasnli}. Lastly, we run qualitative experiments which leverage the relatedness between Jamaican Patois and English to understand which differences between the languages boost or inhibit the effectiveness of cross-lingual transfer.

We hope that JamPatoisNLI prompts long-term research into building NLP tools that consider the particular difficulties and opportunities of NLP for Jamaican Patois and creole languages in general.

\section{Related Work}

\paragraph{Natural Language Inference Datasets.}
Natural language inference (NLI), or recognizing textual entailment, is a standard benchmark task for natural language understanding \cite{rtecooper, rtedagan, nlu}.

The input to the task is a pair of sentences: the premise and the hypothesis. The goal is to output a label -- entailment, neutral or contradiction -- to describe the relationship between the pair. Various approaches have been used to create NLI corpora. The Stanford NLI (SNLI) \cite{snli}, Multi-NLI (MNLI) \cite{williams-etal-2018-broad} and Adversarial NLI (ANLI) \cite{anli} English datasets, esXNLI Spanish dataset \cite{artifacts} Original Chinese Natural Language Inference (OCNLI) dataset \cite{hu-etal-2020-ocnli} and code-mixed Hindi-English dataset \cite{codehindi} all consist of a mixture of pre-existing sentences and crowdsourced sentences. In the Japanese Realistic Textual Entailment Corpus, a collection of pre-existing sentences are filtered and paired using machine learning methods then manually annotated with labels \cite{japannli}.

Other NLI corpora have been made using translation techniques. The Natural Language Inference in Turkish (NLI-TR) dataset \cite{nli-tr} was created using Amazon Translate on SNLI and MNLI. 
The Cross-Lingual NLI (XNLI) Corpus  \cite{xnli} was created by collecting and crowd-sourcing 750 examples then hiring human translators to translate the sentences into 15 languages. Extensions of this dataset to low-resource languages such as AmericasNLI \cite{americasnli} and IndicXNLI \cite{Aggarwal2022IndicXNLIEM} have been created using human and machine translation methods. However, subsequent research has found that translation-based approaches to creating datasets can introduce subtle artifacts which can lead to skewed accuracies for cross-lingual transfer methods \cite{artifacts}. JamPatoisNLI mitigates this problem by using original rather than translated examples.

In spite of the examples given above, generally, there is a relative dearth of datasets and research into methods for low-resource languages across NLI and other tasks. Low-resource languages can be defined as those which are `less studied, resource scarce, less computerized, less privileged, less commonly taught or low density' \cite{lrls}.

\paragraph{Creole Languages in NLP.}

Creole languages are typically low-resource. These languages arise through the process of \textit{creolization} of another class of languages called pidgins. Pidgins emerge as a result of contact between two or more groups of speakers which do not have a common language. A pidgin evolves to become a creole when it becomes the native language of the children of its speakers  \cite{pidgin}.\footnote{We discuss the process of creolization for Jamaican Patois further in Section 3.}

Within the NLP community, a few datasets for different tasks have been created for creoles using a variety of methods.
NaijaSenti is a Twitter human-annotated sentiment analysis dataset which is partly comprised of 14,000 tweets in Nigerian-Pidgin or Naija, which is an English-based creole \cite{naijasenti}.
The authors find that code-switching between these languages and English is a common feature in the dataset.
They explore language adaptive finetuning and zero-shot cross lingual transfer from multilingual pretrained models, and achieve promising results.  Cross-lingual Choice of Plausible Alternatives (XCOPA) \cite{xcopa} is a multilingual dataset for causal common sense reasoning in 11 languages, one of which is Haitian Creole, that was created by translating English COPA. The authors find that across the languages in the dataset, translation based-approaches outperform methods which employ multilingual pretraining and finetuning. A part-of-speech tagging and dependency parsing corpus for Colloquial Singaporean English (Singlish), an English-based creole, has also been created  \cite{singlish1} and further expanded \cite{singlish2} using the Universal Dependencies \cite{ud} scheme. The dataset was created by crawling pages on online Singaporean forums.

Other work has also explored using machine learning methods for identifying and generating creole text. \citet{pidgingen} use contrastive learning to finetune BART \cite{bart} so that the model produces novel dialogue texts in Naija and Yaounde (both English-based creoles). \citet{guadeloupeclassifier} uses a FastText \cite{fasttext} based supervised classifier to identify instances of sentences in Guadeloupean Creole within a multilingual dataset.

The use of machine learning models on creole languages has also been investigated. \citet{creolesoverview} find that standard language models work better than distributionally robust ones on creoles, which shows that these languages are relatively stable. \citet{lent-etal-2022-ancestor}  show that ancestor-to-creole transfer is non-trivial.

\section{Jamaican Patois}

\subsection{Description of the Language}
Jamaican Patois (or Jamaican Creole) is an English-based creole spoken by over 3 million inhabitants on the island and by Jamaicans across the diaspora globally \cite{diaspora}. Jamaican Patois resulted from contact between enslaved Africans brought to the island in the 17th century and British colonists. Because it is a hybrid of the languages spoken by the two groups of people that came in contact, it exists on a continuum that ranges from more dissimilar to less dissimilar to English \cite{davidson1995semantic}.
The terms for the classes in the continuum are the acrolect (variations which are closest to English), the basilect (variations which are furthest from English) and the mesolect (variations which are in between) \cite{inbookcreole} 

Examples of each are shown in Table \ref{continuum}.

\begin{table}[h!t]
\centering
\small
\begin{tabular}{lll}
\toprule
\textbf{Class} & \textbf{Example}\\
\midrule
Basilect & Me a nyam di bickle weh dem gi mi.  \\
Mesolect & Me a eat di food weh dem gi mi.  \\
Acrolect & I'm eating the food that they gave me.   \\
\bottomrule

\end{tabular}
\caption{\label{continuum}
Different translations of `I'm eating the food that they gave me' in Jamaican Patois. The basilectal extreme of the continuum consists of words that are nearly exclusively non-English. On the acrolectal extreme of the spectrum (or Jamaican Standard English), the example is identical to English.}
\end{table}

\subsection{Relevant Linguistic Features}
\paragraph{Unstandardized Orthography.}

Jamaican Patois is primarily a spoken language. Though there have been efforts to develop a formal writing system for the language, none that have been developed are widely used by speakers of Patois. 

Instead, speakers use spelling patterns that reflect how words in Patois are pronounced. This is illustrated in Table \ref{orthography}. In the table, \textit{`I want'} is spelt both \textit{`Me wah'} and \textit{`Mi waa'}: though the phrases yield similar pronunciations, different spellings are used.

\begin{table}[h!]
\centering
\small
\begin{tabular}{lll}
\toprule
\textbf{Jamaican Patois} & \textbf{English}\\
\midrule
Me wah bawl. & I want to cry.  \\
Mi waa cook. & I want to cook.\\
\bottomrule
\end{tabular}
\caption{\label{orthography}
Example of varied spelling of Patois words present in the dataset. }
\end{table}

\paragraph{Vocabulary Overlap with English.}

Since Jamaican Patois is English-based, there is a high degree of overlap between the vocabularies used by the two languages, in spite of differences in spelling, tense and structure. 

We present an example of this in the quote below. Strictly non-English vocabulary (including words such as `a' that have different meanings in English) which are highlighted in bold, account for less than one-third of the words in the sentence.

\begin{quote}
\small
\tt It look like more tourist start come since \textbf{dem} loosen up \textbf{di} restrictions \textbf{dem}. \textbf{Mi} frighten \textbf{fi} see how \textbf{di} beach full \textbf{wen} \textbf{mi} go \textbf{a} Negril \textbf{weh} day.
\end{quote}

\noindent
Therefore, JamPatoisNLI will be useful for evaluating the efficacy of methods for linguistic transfer in scenarios where there is a high degree of overlap between the source and target language.

\paragraph{Negation.}

Common markers of negation used in Jamaican Patois and their English equivalents which feature in the dataset are presented in Table \ref{negmarkers}. Examples of these markers in the dataset are presented in Table \ref{negationeg} in the Appendix.

Negation markers are important linguistic features in the context of NLI datasets, as their presence and interaction with other sentence components are highly relevant to the determination of the right  classification for a given textual entailment  example \cite{negationimpt}.

\begin{table}[h!]
\centering
\small
\begin{tabular}{ll}
\toprule
\textbf{Jamaican Patois} & \textbf{English} \\
\midrule
nuh                      & not/don't/doesn't             \\
cyaa/cyaan               & can't            \\
neva                     & never           \\
\bottomrule
\end{tabular}
\caption{\label{negmarkers}
Markers of negation in Jamaican Patois.}

\end{table}

\begin{table*}[t]
\small
\centering
\begin{tabular}{p{0.4\linewidth} p{0.1\linewidth}p{0.4\linewidth}}
\toprule
\textbf{Premise}                                                     & \textbf{Label}                                                      & \textbf{Hypothesis}                                \\
\midrule
I decided that Christmas haffi ketch me inna good mood!              & {\begin{tabular}[c]{@{}l@{}}\textbf{entailment}\\ E E\end{tabular}}    & Me determined fi happy wen Christmas come!         \\
A dem fi get the money                                               & {\begin{tabular}[c]{@{}l@{}}\textbf{contradiction}\\ C C\end{tabular}} & Dem nuh deserve di money                           \\
mi must make chicken alfredo when mi go home doe                     & {\begin{tabular}[c]{@{}l@{}}\textbf{neutral}\\ N N\end{tabular}}       & mi love fi eat chicken alfredo                     \\
Raisin a get soak in a red label wine fi make cake                   & {\begin{tabular}[c]{@{}l@{}}\textbf{neutral}\\ C N\end{tabular}}       & Mi granny nuh normally mek har cake dem wid raisin \\
I was in juicy beef and yuh know say mi stress out til mi phone drop & {\begin{tabular}[c]{@{}l@{}}\textbf{entailment}\\ E E\end{tabular}}    & Mi phone drop wen mi did deh inna juicy beef  \\ 
\bottomrule
\end{tabular}
\caption{\label{examples}Random sample selected from the 100 double annotated examples in the corpus, with their gold labels and validation labels (abbreviated \texttt{E},  \texttt{N}, \texttt{C}) by each of the annotators.}

\end{table*}

\section{Constructing JamPatoisNLI}

For each example in the dataset, we pulled the premise from a pre-existing text source. Then, a label was randomly selected and a corresponding hypothesis was written by the first author, who speaks and writes Jamaican Patois fluently. Our methodology mirrors that of both MNLI \cite{williams-etal-2018-broad} and ANLI \cite{anli}.

JamPatoisNLI consists of 650 examples split across training, development and validation. Statistics for the corpus are shown in Table \ref{datastats}. A limited availability of native speakers to construct and annotate a large number of examples is a current problem in low-resource NLP \cite{lrls}. 
However, for the purposes of our experiments, the sizes of the training, validation and testing sets are sufficient for exploring few-shot finetuning techniques and obtaining useful signals about the effectiveness of different methods.

\begin{table}[h!]
\centering
\small
\begin{tabular}{lrrrr}
\toprule
\textbf{Statistic}        & \multicolumn{1}{l}{\textbf{Ent.}} & \multicolumn{1}{l}{\textbf{Neu.}} & \multicolumn{1}{l}{\textbf{Con.}} & \multicolumn{1}{l}{\textbf{Total}} \\
\midrule
\#Train          & 84                                      & 83                                   & 83                                         & \textbf{250}              \\
\#Dev            & 66                                      & 67                                   & 67                                         & \textbf{200}              \\
\#Test           & 67                                      & 66                                   & 67                                         & \textbf{200}              \\
\midrule
Avg. Premise Length    & 12.2                                    & 13.6                                 & 11.8                                       & \textbf{12.5}             \\
Avg. Hypothesis Length  & 10.3                                    & 11.9                                 & 10.7                                       & \textbf{11.0}             \\
\#Distinct Words & 1210                                    & 1401                                 & 1187                                       & \textbf{2612}   \\
\bottomrule
\end{tabular}
\caption{\label{datastats}Statistics across the 650 examples in the dataset, by class and in aggregate.}

\end{table}

\subsection{Premise Collection}

Since Jamaican Patois is primarily a spoken language, there is a limited number of textual sources of Patois that are readily available online. However, Patois speakers regularly use the language for communication on social media, and in literature. These are the sources that were used for the premises in the dataset. Around 97\% of examples are drawn from Twitter and the remaining examples are drawn from a cultural website, \href{jamaicans.com}{jamaicans.com}, and from literature by Jamaican poets, Dr. Louise Bennett-Coverley and Shelley Sykes-Coley. The number of examples per source is outlined in Table \ref{sources} in the Appendix.

This method of construction also makes the dataset less prone to effects from translation artifacts which can skew the effectiveness of different cross-lingual transfer techniques. \citet{artifacts} find that when the test dataset is made using translated examples, there is a slight overestimation of the cross-lingual transfer gap as well as the efficacy of the \texttt{TRANSLATE-TRAIN}\footnote{The \texttt{TRANSLATE-TRAIN} technique involves translating the training dataset to the target language.} technique, and an underestimation of the efficacy of the \texttt{TRANSLATE-TEST}\footnote{The \texttt{TRANSLATE-TEST} technique involves translating the testing dataset to the source language.} technique. None of these effects are present when the test dataset is composed of original examples which were not created through translation. Additionally, because the premises of JamPatoisNLI are drawn from natural occurrences of Jamaican Patois written by various speakers of the language, the dataset better reflects the natural writing patterns of speakers than those created using machine or human translation techniques.

\subsection{Hypothesis Construction}

The set of hypotheses in the corpus is comprised of novel sentences constructed by our first author, who is a native speaker of Jamaican Patois. For each premise, a corresponding hypothesis was written so that the pair's classification would be either \texttt{entailment}, \texttt{neutral} or \texttt{contradiction}. The criteria used for assignment of pairs to each class is shown in Figure \ref{fig:criteria} in the Appendix.

The constructed hypothesis in each example mimics the diverse spelling conventions and writing patterns used in the corresponding pre-existing premise. As such, the non-standardized nature of Jamaican Patois is reflected in both the collected and constructed sentences in the dataset.

 In order to maximize the linguistic diversity of examples in the dataset, each premise was used to generate a single hypothesis (rather than three hypotheses generated per premise, which was done for MNLI \cite{williams-etal-2018-broad}).

\subsection{Label Validation}

A random sample of 100 sentence pairs evenly distributed across the three classes was double annotated by fluent speakers of Jamaican Patois. We recruited volunteer annotators by reaching out to friends and colleagues. The labelling criteria given to the annotators were the same as those used to generate the hypotheses, and are outlined in Appendix Figure \ref{fig:criteria}. In Table \ref{annotator}, we present statistics for inter-annotator agreement for these examples. The Fleiss Kappa accuracy for the dataset was 88.99\% while the percentage accuracy was 89.00\%.

\begin{table}[t]
\small
\begin{tabular}{lrr}
\toprule
\textbf{Metric}           & \multicolumn{1}{l}{\textbf{Accuracy}} & \multicolumn{1}{l}{\textbf{Counts}} \\
\midrule
Fleiss K                  & 88.99\%                               & 100                                 \\
\% Accuracy               & 89.00\%                               & 100                                 \\
Neutral \% Accuracy       & 75.76\%                               & 33                                  \\
Entailment \% Accuracy    & 100.00\%                              & 34                                  \\
Contradiction \% Accuracy & 90.91\%                               & 33  \\        
\bottomrule
\end{tabular}
\caption{\label{annotator}
Inter-annotator agreement.  We count a classification as accurate if both annotators agreed with the original annotations in the dataset.}

\end{table}
\section{Experiments and Results}

Across our experiments, our goals are to:
\begin{enumerate}
    \item Provide benchmarks for JamPatoisNLI thus determining the difficulty of the dataset and effectiveness of cross-lingual transfer.
    \item  Compare the effectiveness of cross-lingual transfer on JamPatoisNLI (a language that is \textit{related} to language(s) present in the training corpus of each of the pretrained models we examine), to cross-lingual transfer on AmericasNLI  (which contains languages that are \textit{unrelated} to any language(s) present in the training corpus of each pretrained model).
     \item Leverage the nature of Jamaican Patois as a creole to further understand cross-lingual transfer.
\end{enumerate}
 The experiments that we conduct are done in the zero-shot and few-shot settings.
 
\subsection{General Setup}

In our experiments, we use English BERT, multilingual BERT \cite{mbert}, English RoBERTa \cite{roberta} and XLM-RoBERTa \cite{xlm-r} as our base pretrained models. We use a two-layer perceptron with ReLU activations for the classification head, and first finetune on the MNLI training dataset. We use cased and uncased versions of each BERT-based pretrained model, and experiment with frozen and unfrozen versions,\footnote{In our frozen model, all parameters of the pretrained base models are fixed during finetuning so that only the NLI classification head is updated, while for our unfrozen models, all model parameters are allowed to update.} for a total of eight types of BERT-based models. For our RoBERTa-based models, we also experiment with frozen and unfrozen versions for a total of four types of RoBERTa-based models. Throughout our experiments with the twelve model types, we make comparisons among the BERT-based models and the RoBERTa-based models separately.

 To select the twelve MNLI finetuned models that we use for our few-shot experiments, we conduct a hyperparameter search over dropouts in the range [0.2, 0.5], batch sizes in the range [8, 32], learning rates in the range [1e-05, 1e-06] and epoch counts in the range [2, 10] and pick those that achieved reasonable accuracies on the MNLI development dataset (above 86\% for unfrozen models and above 62\% for frozen models).

Among the twelve selected models finetuned on MNLI, we evaluate the zero-shot and few-shot performance on each of our target datasets to determine which model types produce the highest accuracy. To compare the types of models, we fix the hyperparameters to the values in Table \ref{hyperparameters} in the Appendix, and average over three experiments with different seeds. Then, from among the eight finetuned BERT-based models, we pick the type that achieved the highest scores for the maximum number of few-shot training examples for each our validation datasets (JamPatoisNLI and AmericasNLI). We also do the same for the four finetuned RoBERTa-based models.

 After we select the best out of the model types among the models finetuned on MNLI and further finetuned on the target fewshot datasets, we perform a final hyperparameter sweep.  Tables \ref{jamfewshothyper} and \ref{xlmjamfewshothyper} show the final set of hyperparameters that we arrived at after we conducted our sweep for the best models on the JamPatoisNLI and AmericasNLI validation sets among our BERT-based models and RoBERTa-based models.

In our few-shot finetuning setup, we select one example from each class for each ``shot''. For instance, using this convention, two-shot finetuning involves finetuning using six examples in total: two from each of the three NLI classes. Additionally, during few-shot finetuning, we keep all layers of the base model unfrozen.

\subsection{Benchmarks for JamPatoisNLI}

\begin{table}[t]
\centering
\small
\begin{tabular}{@{}lrrrr@{}}
\toprule
\textbf{Hyperparameter} & \textbf{\begin{tabular}[c]{@{}l@{}}Best Model on\\ JamPatoisNLI\end{tabular}} &
\textbf{\begin{tabular}[c]{@{}l@{}}Best Model on \\ AmericasNLI  \end{tabular}} \\
\midrule
Finetune epoch ct.   & 5   & 5                                                 \\
Finetune batch size    & 16     & 16                                              \\
Finetune learning rate & 1e-05      & 1e-05                                          \\
Finetune dropout       & 0.3               & 0.3                                   \\
Few shot \# of iter.     & 200               & 100                 \\
Few shot batch size           & 16              & 8                   \\
Few shot learning rate        & 5e-05           & 1e-05                   \\
Few shot dropout              & 0.25                   & 0.25            \\
\bottomrule
\end{tabular}
\caption{\label{jamfewshothyper}
Final hyperparameters for best BERT-based model on JamPatoisNLI (\texttt{bert-uncased-unfrozen}) and AmericasNLI (\texttt{mbert-cased-unfrozen}).}

\end{table}

\begin{table}[t]
\centering
\small
\begin{tabular}{@{}lrrrr@{}}
\toprule
\textbf{Hyperparameter} & \textbf{\begin{tabular}[c]{@{}l@{}}Best Model on \\ JamPatoisNLI\end{tabular}} &
\textbf{\begin{tabular}[c]{@{}l@{}}Best Model on\\ AmericasNLI \end{tabular}} \\
\midrule
Finetune epoch ct.   & 3   & 5                                                 \\
Finetune batch size    & 32     & 16                                              \\
Finetune learning rate & 1e-05      & 1e-05                                          \\
Finetune dropout       & 0.2               & 0.3                                   \\
Few shot \# of iter.     & 200               & 100                 \\
Few shot batch size           & 16              & 16                   \\
Few shot learning rate        & 1e-05           & 1e-05                   \\
Few shot dropout              & 0.25                   & 0.25            \\
\bottomrule
\end{tabular}
\caption{\label{xlmjamfewshothyper}
Final hyperparameters for best RoBERTa-based model on JamPatoisNLI (\texttt{roberta-unfrozen}) and AmericasNLI (\texttt{xlm-unfrozen}).}

\end{table}
\begin{table*}[]
\small
\centering
\begin{tabular}{rrrrrrr}
\toprule
\multicolumn{1}{l}{\textbf{\begin{tabular}[c]{@{}l@{}}\# of Fewshot\\ Class Triples\end{tabular}}} & \multicolumn{1}{l}{\textbf{Maj. Base.}} & \multicolumn{1}{l}{\textbf{\begin{tabular}[c]{@{}l@{}}Hyp. Only Base.\\ (bert-uncased-\\ unfrozen)\end{tabular}}} & \multicolumn{1}{l}{\textbf{\begin{tabular}[c]{@{}l@{}}bert-uncased-\\ unfrozen\end{tabular}}} & \multicolumn{1}{l}{\textbf{\begin{tabular}[c]{@{}l@{}}mbert-uncased-\\ unfrozen\end{tabular}}} & \multicolumn{1}{l}{\textbf{\begin{tabular}[c]{@{}l@{}}roberta-\\ unfrozen\end{tabular}}} & \multicolumn{1}{l}{\textbf{xlm-unfrozen}} \\

\midrule
0                                                                                                  & 33.50                                   & 38.50                                                                                                             & 56.00                                                                                         & 50.00                                                                                          & 67.50                                         & 56.00                                     \\
1                                                                                                  & 33.50                                   & 38.17                                                                                                             & 54.50                                                                                         & 52.17                                                                                          & 68.17                                         & 57.50                                     \\
2                                                                                                  & 33.50                                   & 37.17                                                                                                             & 56.83                                                                                         & 53.33                                                                                          & 69.17                                         & 58.17                                     \\
4                                                                                                  & 33.50                                   & 37.00                                                                                                             & 51.00                                                                                         & 52.33                                                                                          & 66.83                                         & 57.67                                     \\
8                                                                                                  & 33.50                                   & 35.83                                                                                                             & 52.17                                                                                         & 51.17                                                                                          & 68.83                                         & 57.50                                     \\
16                                                                                                 & 33.50                                   & 38.83                                                                                                             & 56.17                                                                                         & 53.50                                                                                          & 70.17                                         & 58.83                                      \\
32                                                                                                 & 33.50                                   & 38.50                                                                                                             & 61.17                                                                                         & 63.83                                                                                          & 73.00                                         & 70.00                                     \\
64                                                                                                 & 33.50                                   & 46.33                                                                                                             & 64.50                                                                                         & 65.17                                                                                          & 76.33                                         & 72.50                                     \\
83                                                                                                 & 33.50                                   & 43.33                                                                                                             & 66.17                                                                                         & 65.33                                                                                          & 76.50                                         & 75.17     \\   
\bottomrule

\end{tabular}
\caption{\label{patoisnumbers} Zero-shot and few-shot accuracies for different models evaluated on JamPatoisNLI averaged over three experiments with different seeds. The best models were chosen based on results for the validation set.}
\end{table*}

\paragraph{Setup.} For JamPatoisNLI, the best BERT-based model type was the unfrozen uncased English BERT model (\texttt{bert-uncased-unfrozen}) based on accuracies on the validation set. Using the hyperparameters in Table \ref{jamfewshothyper}, we also make comparisons to a hypothesis only baseline (\texttt{bert-uncased-unfrozen}), as well as the best multilingual BERT-based model on JamPatoisNLI, which was the unfrozen uncased multilingual BERT model (\texttt{mbert-uncased-unfrozen}).

The best RoBERTa-based model type was the unfrozen English RoBERTa model (\texttt{roberta-unfrozen}). We also include results for the best multilingual RoBERTa-based model on the dataset, which was the unfrozen XLM-RoBERTa model (\texttt{xlm-unfrozen}). The hyperparameters that we used are listed in Table \ref{xlmjamfewshothyper}.

\paragraph{Results.}  Our results on the test set are presented in Table \ref{patoisnumbers}. We found that with the maximum number training of examples, \texttt{bert-uncased-unfrozen} and \texttt{mbert-uncased-unfrozen} had relatively similar accuracies when all few-shot examples were used (66.17\% and 65.33\% respectively). We also found that \texttt{roberta-unfrozen}  and \texttt{xlm-unfrozen} achieve similar accuracies on the full fewshot dataset (76.50\% and 75.17\%) respectively.

The two RoBERTa-based models significantly outperformed the two BERT-based models -- in fact, the zero-shot accuracy on the \texttt{roberta-unfrozen} model (67.50\%) outperforms both BERT based models when they are finetuned on the full few-shot dataset.

For our best model (\texttt{xlm-unfrozen}), the standard deviation in percentage accuracy for the maximum number of few-shot examples across ten experiments was 0.75\% when evaluated on the validation set and 1.43\% when evaluated on the test set.

\subsection{Comparisons with AmericasNLI}

\paragraph{Setup.} A natural comparison point for JamPatoisNLI is AmericasNLI \cite{americasnli} as it is also a low-resource NLI dataset. However, unlike Jamaican Patois, the languages in the corpus are not closely related to any high-resource languages for which there are large pretrained language models or large natural language inference training datasets. In particular, the languages in AmericasNLI do not belong to the same family as any of the languages in the two most commonly used multilingual pretrained language models -- multilingual BERT \cite{mbert} and XLM-R \cite{xlmr}.
  JamPatoisNLI is \textit{unseen} from the perspective of existing pretrained monolingual or multilingual models but \textit{related} to the source language(s) involved in transfer learning, whereas AmericasNLI is both \textit{unseen} and \textit{unrelated}.

  For our experiments, we use five of the languages in the AmericasNLI dataset, and create a randomly selected 250-200-200 train-dev-test split from among the examples in the original development dataset for each language (shown in Table \ref{americasnli} in the Appendix) to mirror the number of examples present in each of the splits in JamPatoisNLI.

For the AmericasNLI languages, the best BERT-based model type based on results on the validation set was the unfrozen cased multilingual BERT model (\texttt{mbert-cased-unfrozen}). The best RoBERTa-based model type was the unfrozen XLM-RoBERTa model (\texttt{xlm-unfrozen}).   

\begin{table}[]
\small
\centering
\begin{tabular}{rrrrr}
\toprule
\multicolumn{1}{l}{\textbf{\begin{tabular}[c]{@{}l@{}}\  \end{tabular}}} & \multicolumn{2}{l}{\textbf{\begin{tabular}[c]{@{}l@{}}Avg. AmericasNLI \\ Accuracy\end{tabular}}}                                                                                   & \multicolumn{2}{l}{\textbf{\begin{tabular}[c]{@{}l@{}}Patois\\ Accuracy\end{tabular}}}                                                                                                                                                \\
\midrule
\textbf{ Num.}                                                                                               & \multicolumn{1}{l}{\textbf{\begin{tabular}[c]{@{}l@{}}mbert-\\ cased-\\ unfrozen\end{tabular}}} & \multicolumn{1}{l}{\textbf{\begin{tabular}[c]{@{}l@{}}xlm-\\ unfrozen\end{tabular}}} & \multicolumn{1}{l}{\textbf{\begin{tabular}[c]{@{}l@{}}bert-\\ uncased-\\ unfrozen\end{tabular}}} & \multicolumn{1}{l}{\textbf{\begin{tabular}[c]{@{}l@{}}roberta-\\ unfrozen\end{tabular}}} \\
\midrule

0                                                                                                        & 42.00                                                                                           & 39.60                                                                                & 56.00                                                                                            & 67.50                                                                                    \\
1                                                                                                        & 41.83                                                                                           & 39.17                                                                                & 54.50                                                                                            & 68.17                                                                                    \\
2                                                                                                        & 42.67                                                                                           & 39.50                                                                                & 56.83                                                                                            & 69.17                                                                                    \\
4                                                                                                        & 42.67                                                                                           & 40.03                                                                                & 51.00                                                                                            & 66.83                                                                                    \\
8                                                                                                        & 42.70                                                                                           & 39.93                                                                                & 52.17                                                                                            & 68.83                                                                                    \\
16                                                                                                       & 43.63                                                                                           & 42.77                                                                                & 56.17                                                                                            & 70.17                                                                                    \\
32                                                                                                       & 46.40                                                                                           & 46.07                                                                                & 61.17                                                                                            & 73.00                                                                                    \\
64                                                                                                       & 48.87                                                                                           & 47.40                                                                                & 64.50                                                                                            & 76.33                                                                                    \\
83                                                                                                       & 49.23                                                                                           & 48.83                                                                                & 66.17                                                                                            & 76.50      \\                                        \bottomrule
                                     
\end{tabular} \caption{\label{bestpatoistable}Test set accuracies for best BERT-based and RoBERTa-based models on the JamPatoisNLI dataset \texttt{(bert-uncased-unfrozen, roberta-unfrozen)} and on the AmericasNLI  dataset \texttt{(mbert-cased-unfrozen, xlm-unfrozen)}. Experiments are averaged over three seeds and the best models were chosen based on results for the validation set.}

\end{table}
\begin{figure}[t]

\includegraphics[width=1\linewidth]{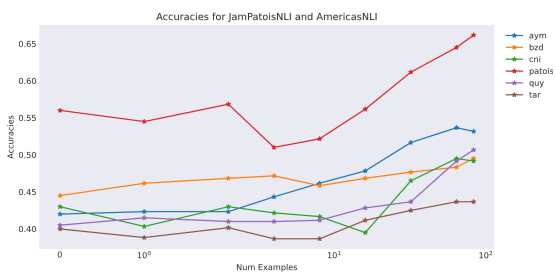}
\caption{\label{figureplot2} Plots for the best AmericasNLI model (\texttt{mbert-cased-unfrozen}) on each language, and the best JamPatoisNLI model (\texttt{bert-uncased-unfrozen}). Experiments are averaged over three seeds and the best models were chosen based on results for the val. set.}
\end{figure}

\begin{figure}[t]

\includegraphics[width=1\linewidth]{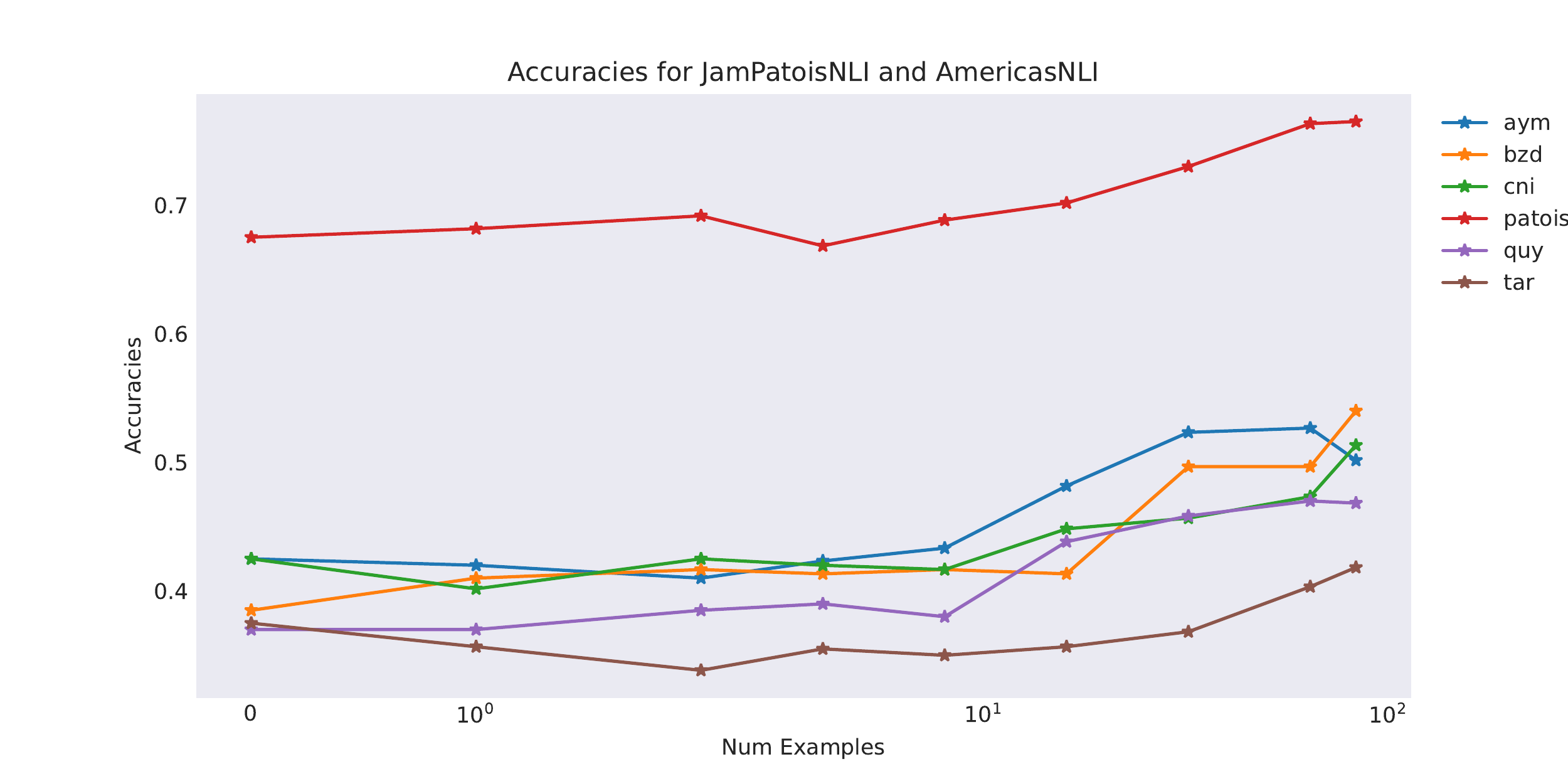}
\caption{\label{figureplot3} Plots for the best AmericasNLI model (\texttt{xlm-unfrozen}) on each language, and the best JamPatoisNLI model (\texttt{roberta-unfrozen}). Experiments are averaged over three seeds and the best models were chosen based on results for the val. set.}
\end{figure}
\paragraph{Results.} We present the results of our experiments on the test set in Table \ref{bestpatoistable}. We found that there was a significant gap in accuracies on JamPatoisNLI and AmericasNLI. Across all experiments, both zero-shot and few-shot accuracies for the JamPatoisNLI dataset exceeded those for the AmericasNLI dataset. The best JamPatoisNLI model achieved a zero-shot accuracy of 67.50\% while the best AmericasNLI model achieved a zero-shot accuracy of 42.00\% (both compared to a 33.50\% majority baseline).

This shows that the language relatedness between Jamaican Patois and English significantly boosts the effectiveness of cross-lingual transfer learning even in the zero-shot case. 
For the few-shot setting, the highest  accuracy achieved on the JamPatoisNLI dataset was 76.50\%. The highest average accuracy achieved on the AmericasNLI dataset was 49.23\%.

  The plots comparing the best JamPatoisNLI model to the best AmericasNLI model on each of the respective datasets for BERT-based models and RoBERTa-based models are shown in Figures \ref{figureplot2} and \ref{figureplot3}. For the BERT-based models, we see that cross-lingual transfer augmented by few-shot learning is quite effective for JamPatoisNLI, whereas the gains for AmericasNLI languages are rather modest. Tabulated results for these experiments can be found in Appendix Tables \ref{accuraciesbestamerica} and \ref{xlmaccuraciesbestamerica}.

\subsection{Experiments with Transitioning from Jamaican Patois to English}

\begin{table*}[t]
\centering
\small
\begin{tabular}{lllll}
\toprule
\textbf{Change} & \textbf{Premise}                                                                                          & \textbf{Hypothesis}                                                                                           & \textbf{Tgt.} & \textbf{M1 M2 M3}  \\
\midrule
- & \begin{tabular}[c]{@{}l@{}}Any day \textcolor{red}{\textbf{mi}} master \\ LumaFusion,\\ mi lef my work.\end{tabular}                  & \begin{tabular}[c]{@{}l@{}}As soon as \textcolor{red}{\textbf{mi}} good \\wid LumaFusion, \\mi a quit mi job\end{tabular}                & E               & C \quad C \quad C           \\
\midrule

\begin{tabular}[c]{@{}l@{}} \textbf{Pronoun:} \\   mi $\rightarrow$ I \end{tabular}  
 & \begin{tabular}[c]{@{}l@{}}Any day \textcolor{red}{\textbf{I}} master \\ LumaFusion, \\mi \textcolor{orange}{\textbf{lef}} my work\end{tabular}                    & \begin{tabular}[c]{@{}l@{}}As soon as \textcolor{red}{\textbf{I'm}} good \\ wid  LumaFusion,\\ mi \textcolor{orange}{\textbf{a quit}} my job\end{tabular}               & E               & C \quad C \quad C              \\
 \midrule
 \begin{tabular}[c]{@{}l@{}}\textbf{Verb:} \\lef/quit $\rightarrow $ leaving/quitting \end{tabular}
  & \begin{tabular}[c]{@{}l@{}}Any day I master \\ LumaFusion, \textcolor{brown}{\textbf{mi}}\\ \textcolor{orange}{\textbf{leaving}} my work\end{tabular}                & \begin{tabular}[c]{@{}l@{}}As soon as I'm good \\wid LumaFusion,\\ \textcolor{brown}{\textbf{mi}} \textcolor{orange}{\textbf{quitting}} this job\end{tabular}           & E               & \textbf{E} \quad \textbf{E}  \quad \textbf{E}  \\
  \midrule
\begin{tabular}[c]{@{}l@{}} \textbf{Pronoun:} \\   mi $\rightarrow$ I \end{tabular}  & \begin{tabular}[c]{@{}l@{}}\textcolor{violet}{\textbf{Any day}} I master \\ LumaFusion, \textcolor{brown}{\textbf{I'm}}\\ leaving my job\end{tabular}                & \begin{tabular}[c]{@{}l@{}}As soon as I'm good\\ \textcolor{violet}{\textbf{wid}} LumaFusion,\\ \textcolor{brown}{\textbf{I'm}} quitting my job.\end{tabular}           & E               &\textbf{E} \quad \textbf{E}  \quad \textbf{E} \\
\midrule
\begin{tabular}[c]{@{}l@{}} \textbf{Determiner/Preposition:} \\   Any day/wid $\rightarrow$ The day that/with \end{tabular}  & \textbf{\begin{tabular}[c]{@{}l@{}}\textcolor{violet}{\underline{The day that}} I master \\ LumaFusion, I'm\\ leaving my job.\end{tabular}} & \textbf{\begin{tabular}[c]{@{}l@{}}As soon as I'm good\\ \textcolor{violet}{\underline{with}} LumaFusion, \\I'm quitting my job.\end{tabular}} & \textbf{E}      & \textbf{E} \quad \textbf{E}  \quad \textbf{E} \\
\bottomrule
\end{tabular}
\caption{\label{transition}
Sample from Jamaican Patois to English transition dataset. The final example is in English, and we present predictions made by three models finetuned with our Patois few-shot training dataset using the parameters for the best JamPatoisNLI model in Table \ref{jamfewshothyper}.}. 

\end{table*}

\paragraph{Setup.} A key characteristic of Jamaican Patois is that it exists on a spectrum that ranges from highly dissimilar to English (the basilect), to highly similar to English (the acrolect). We experiment with 83-shot classification (the full set of examples in our few-shot training dataset) on an augmented test dataset derived from pairs that were incorrectly classified by at least two of the three models in our original few-shot experiments. To construct this dataset, we picked a single example for each type of  misclassification with respect to the three NLI labels, for a total of 6 examples from the original dataset  (which mostly fell on various points on the mesolectal range of the creole spectrum). We then wrote English translations for each of these examples (which would fall on the acrolectal end of the creole spectrum) and hand-wrote intermediate translations between them that are all valid Jamaican Patois to qualitatively study whether (and for what changes) along the path the label becomes correct. We conduct few-shot finetuning using our original training set for three models with different seeds using the parameters for the best BERT-based JamPatoisNLI model (\texttt{bert-uncased-unfrozen}), listed in Table \ref{jamfewshothyper}.

\paragraph{Results.} We present a qualitative example of this experiment in Table \ref{transition}. Here,  changing the verb from Jamaican Patois to English caused the models to switch to the correct classification. The three models switched to the correct prediction for a change prior to the full translation of the Jamaican Patois example to English for all but one of the originally misclassified examples in our experiments.

\section{Discussion}

We see that the relatedness between Jamaican Patois and English strongly contributes to the effectiveness of cross-lingual transfer in both zero-shot and few-shot settings. 
Additionally, although natural language inference is a higher order reasoning task, our models achieved relatively high accuracy on the JamPatoisNLI dataset by learning the task from MNLI examples in English.

A natural question that arises based on these results, is whether vocabulary overlap is the primary factor that led to the boost in effectiveness of transfer learning in these experiments, or whether a higher order notion of similarity is a larger factor. Comparing zero-shot and few-shot accuracies for other languages that are closely related to English but do not share the same degree of vocabulary overlap as an English-based creole (such as German) might be an interesting line of future research.

Interestingly, though Jamaican Patois developed as a result of contact between speakers of English and speakers of West African languages (some of which are present in multilingual BERT's and XLM-RoBERTa's training corpus), the multilingual models were not more effective base pretrained language models than the monolingual models.  Another possible direction for future research might be to determine whether there are methods that allow for more effective leveraging of the multilingual characteristic of the models during finetuning for creole target languages.

\section{Conclusion}

JamPatoisNLI is a natural language inference dataset in an English-based creole, constructed from existing and novel examples of Jamaican Patois. Our experiments show that the language's relatedness to English significantly boosts the effectiveness of cross-lingual transfer, even for the higher order task of natural language inference in both zero-shot and few-shot settings. We hope that the creation of this dataset encourages further research in the field on methods to improve cross-lingual transfer for creole target languages, and the creation of other low-resource language and creole language datasets. 

\section*{Acknowledgements}  We thank
Roxanne Dobson, Ghawayne Calvin, Danielle
Roberts, Khaesha Brooks, Dominique Lyew and
Ana-Katrina Donaldson for volunteering to be dataset annotators.
We also thank Prof. Christopher Potts
(cgpotts@stanford.edu) for comments on the paper.
RA was partially supported by a Siebel Scholarship, a Google Generation Scholarship and the SWE Motorola Solutions Foundation Engineering Scholarship. JH was supported by an NSF Graduate Research Fellowship under grant number DGE-1656518.

\section{Limitations}

One limitation of our research is related to the fact that Jamaican Patois is a low-resource language. The size of the dataset splits (particularly, the validation and test sets) are much smaller than those of high-resource language datasets.

Further, the differences observed between the AmericasNLI and JamPatoisNLI datasets are not necessarily solely due to differences in language similarity to the source languages: another contributing factor might be differences in difficulty for the two datasets. 

\bibliography{custom}
\bibliographystyle{acl_natbib}
\newpage
\appendix

\section{Appendix}
\label{sec:appendix}

\subsection{Finetuning with BitFit}

BitFit is a sparse parameter efficient finetuning method introduced for use with small-to-medium sized training datasets which involves finetuning only the bias terms of a pretrained language model \cite{bitfit}. As an initial approach for few-shot finetuning, we experimented with using BitFit using the same hyperparameters described in our prior experiments (in Table \ref{jamfewshothyper}) for the best JamPatoisNLI model (English BERT uncased unfrozen), but increasing the learning rate by one order of magnitude as the authors do in the paper to 5e-04.

In Table \ref{bitfit}, we present the results for few-shot finetuning using the BitFit method \cite{bitfit} in comparison with the vanilla finetuning method (in which all model parameters are left unfrozen). In the zero-shot setting and in the cases where there are a small number of few-shot examples, the two techniques perform similarly, but BitFit begins to underperform relative to the vanilla method with more few-shot examples.

\begin{table}[h!]
\centering
\small
\begin{tabular}{lll}
\toprule
\textbf{Num Examples} & \textbf{Jam} & \textbf{Jam-BitFit} \\
\midrule
0                    & 56.00                   & 56.00                \\
1                    & 54.50                   & 55.83                \\
2                    & 56.83                   & 55.67                \\
4                    & 51.00                   & 55.83                \\
8                    & 52.17                   & 55.83                \\
16                   & 56.17                   & 55.83                \\
32                   & 61.17                   & 54.67                \\
64                   & 64.50                   & 58.00                \\
83                   & 66.17                   & 58.67               

          \\
\bottomrule
\end{tabular}
\caption{\label{bitfit} Comparison for zero-shot and few-shot finetuning using BitFit and the vanilla finetuning technique. Experiments are averaged over three seeds, and are reported on the test dataset.}

\end{table}

\begin{table}[h]
\small
\centering
\begin{tabular}{lll}
\toprule
\textbf{Source} & \textbf{Examples}\\
\midrule
Twitter & 634  \\
Anthology: Shelley Sykes-Coley & 6  \\
Poetry: Rt. Hon. Dr. Louise Bennett-Coverley & 4  \\
Online blog  & 6 \\

\bottomrule
\end{tabular}
\caption{\label{sources}
Sources for premises in the dataset.}
\end{table}

\begin{table}[h!]
\small
\begin{tabular}{lllrr}
\toprule
\textbf{Language}  & \textbf{ISO} & \textbf{Family}      & \multicolumn{1}{l}{\textbf{Dev}} & \multicolumn{1}{l}{\textbf{Test}} \\
\midrule
Aymara    & aym & Aymaran     & 743                     & 750                      \\
Asháninka & cni & Arawak      & 658                     & 750                      \\
Bribri    & bzd & Chibchan    & 743                     & 750                      \\
Quechua   & quy & Quechuan    & 743                     & 750                      \\
Rarámuri  & tar & Uto-Aztecan & 743                     & 750                     \\
\bottomrule
\end{tabular}
\caption{\label{americasnli}
Languages used from the AmericasNLI dataset and the sizes of the original splits.}

\end{table}
 \begin{table}[t]
\centering
\small
\begin{tabular}{lr}
\toprule
\textbf{Hyperparameter} & \multicolumn{1}{l}{\textbf{Values}} \\
\midrule
Batch size              & 8, 16                               \\
Learning rate           & 1e-05, 5e-05                        \\
Number of iterations    & 100, 200     \\ 
\bottomrule

\end{tabular}
\caption{\label{fewshothyper}
Values used for few-shot hyperparameter sweep. Experiments are averaged over three seeds.}

\end{table}
\begin{table}[h!]
\centering
\small
\begin{tabular}{lr}
\toprule
\textbf{Hyperparameter} & \multicolumn{1}{l}{\textbf{Value}} \\
\midrule
Batch size              & 8                                  \\
Learning rate           & 1e-05                           \\
Number of iterations    & 100                                \\
Dropout                 & 0.25                              \\
\bottomrule
\end{tabular}
\caption{\label{hyperparameters}
Hyperparameters used for model type selection. Experiments are averaged over three seeds.}

\end{table}
\begin{figure}[h!]
\small
\centering
\noindent\fbox{%
    \parbox{0.8\linewidth}{%
    \textbf{Entailment.} \\
        (a) Given the premise, a reasonable reader would conclude that the hypothesis must also be true.\\
        (b) The hypothesis is necessarily consistent with the premise.\\
        (c) If a speaker holds the sentiment or opinion expressed in premise, then a reasonable reader would conclude that they also hold the sentiment or opinion expressed in hypothesis.\\

 \textbf{Contradiction.}\\
        (a) Given the premise, a reasonable reader would conclude that the hypothesis must be false.\\ (b) The hypothesis is necessarily inconsistent with the premise. \\(c) If a speaker holds the sentiment or opinion expressed in premise, then a reasonable reader would conclude that they do not hold the sentiment or opinion expressed in hypothesis.\\

\textbf{Neutral}\\
        (a) Given the premise, a reasonable reader would conclude that the hypothesis could be either true or false.\\
        (b) The hypothesis is neither necessarily inconsistent nor necessarily consistent with the premise.\\
        (c) If a speaker holds the sentiment or opinion expressed in premise, then a reasonable reader would conclude that it may or may not be true that they hold the sentiment or opinion expressed in hypothesis.

    }%
}
  \caption{Labelling criteria used to generate each hypothesis based on the premise, and given as labelling guidelines to dataset validators.}
\label{fig:criteria}

\end{figure}

\begin{table*}[h!]
\centering
\small
\begin{tabular}{lll}
\toprule
\textbf{Premise}                                                                                                         & \textbf{Hypothesis}                                                                                          & \textbf{Label} \\
\midrule
\begin{tabular}[c]{@{}l@{}}Jason mi deh cook and me nah\\  mek u mek di likkle bickle \\ bun up!\end{tabular}            & \begin{tabular}[c]{@{}l@{}}Jason neva eat cook food\\  from da restaurant deh inna\\  im life\end{tabular}   & neutral        \\
\begin{tabular}[c]{@{}l@{}}And if dem tek everything\\  and all mi have a my breathe ,\\  mi happy same way\end{tabular} & \begin{tabular}[c]{@{}l@{}}Nuh matta weh dem waa\\  tek from mi glad as long as\\  mi have life\end{tabular} & entailment     \\
\begin{tabular}[c]{@{}l@{}}Mi nuh bada waa get married... \\ ever\end{tabular}                                           & Mi cyaa wait fi get married                                                                                  & contradiction \\
\bottomrule
\end{tabular}
\caption{\label{negationeg}
Examples of negation markers in examples from each of the three classes in the dataset.}

\end{table*}

\begin{table*}[t]
\centering
\small
\begin{tabular}{lllllll}
\toprule
\textbf{Num Examples} & \textbf{aym} & \textbf{bzd} & \textbf{cni} & \textbf{quy} & \textbf{tar} & \textbf{jam} \\
\midrule
0                                         & 42.00                            & 44.50                            & 43.00                            & 40.50                            & 40.00                            & 56.00                            \\
1                                         & 42.33                            & 46.17                            & 40.33                            & 41.50                            & 38.83                            & 54.50                            \\
2                                         & 42.33                            & 46.83                            & 43.00                            & 41.00                            & 40.17                            & 56.83                            \\
4                                         & 44.33                            & 47.17                            & 42.17                            & 41.00                            & 38.67                            & 51.00                            \\
8                                         & 46.17                            & 45.83                            & 41.67                            & 41.17                            & 38.67                            & 52.17                            \\
16                                        & 47.83                            & 46.83                            & 39.50                            & 42.83                            & 41.17                            & 56.17                            \\
32                                        & 51.67                            & 47.67                            & 46.50                            & 43.67                            & 42.50                            & 61.17                            \\
64                                        & 53.67                            & 48.33                            & 49.50                            & 49.17                            & 43.67                            & 64.50                            \\
83                                        & 53.17                            & 49.50                            & 49.17                            & 50.67                            & 43.67                            & 66.17                           
       \\
\bottomrule
\end{tabular}
\caption{\label{accuraciesbestamerica}
Zero-shot and few-shot plot for the best BERT-based AmericasNLI model (\texttt{mbert-cased-unfrozen}) accuracies for each language in the dataset and the best BERT-based JamPatoisNLI model (\texttt{bert-uncased-unfrozen}). Experiments are averaged over three seeds and the best models were chosen based on results for the validation set.}
\end{table*}

\begin{table*}[t]
\centering
\small
\begin{tabular}{lllllll}
\toprule
\textbf{Num Examples} & \textbf{aym} & \textbf{bzd} & \textbf{cni} & \textbf{quy} & \textbf{tar} & \textbf{jam} \\
\midrule
0                                         & 42.50                            & 38.50                            & 42.50                            & 37.00                            & 37.50                            & 67.50                            \\
1                                         & 42.00                            & 41.00                            & 40.17                            & 37.00                            & 35.67                            & 68.17                            \\
2                                         & 41.00                            & 41.67                            & 42.50                            & 38.50                            & 33.83                            & 69.17                            \\
4                                         & 42.33                            & 41.33                            & 42.00                            & 39.00                            & 35.50                            & 66.83                            \\
8                                         & 43.33                            & 41.67                            & 41.67                            & 38.00                            & 35.00                            & 68.83                            \\
16                                        & 48.17                            & 41.33                            & 44.83                            & 43.83                            & 35.67                            & 70.17                            \\
32                                        & 52.33                            & 49.67                            & 45.67                            & 45.83                            & 36.83                            & 73.00                            \\
64                                        & 52.67                            & 49.67                            & 47.33                            & 47.00                            & 40.33                            & 76.33                            \\
83                                        & 50.17                            & 54.00                            & 51.33                            & 46.83                            & 41.83                            & 76.50                           

       \\
\bottomrule
\end{tabular}
\caption{\label{xlmaccuraciesbestamerica}
Zero-shot and few-shot plot for the best RoBERTa-based AmericasNLI model (\texttt{xlm-unfrozen}) accuracies for each language in the dataset and the best RoBERTa-based JamPatoisNLI model (\texttt{roberta-unfrozen}). Experiments are averaged over three seeds and the best models were chosen based on results for the validation set.}
\end{table*}

\end{document}